\documentclass{article}
\usepackage{ijcai26}

\usepackage{times}
\usepackage{soul}
\usepackage{url}
\usepackage[hidelinks]{hyperref}
\usepackage[utf8]{inputenc}
\usepackage[small]{caption}
\usepackage{graphicx}
\usepackage{amsmath}
\usepackage{amsthm}
\usepackage{booktabs}
\usepackage{multirow}
\usepackage{algorithm}
\usepackage{algorithmic}
\usepackage[switch]{lineno}
\usepackage{xcolor}
\usepackage{colortbl}
\usepackage{pgfplots}
\pgfplotsset{compat=1.17}

\title{O\textsc{Mind}: Framework for Knowledge Grounded Finetuning and Multi-Turn Dialogue Benchmark for Mental Health LLMs
}

\author{
    Suraj Racha, 
    Prashant Harish Joshi, 
    Utkarsh Maurya, 
    Nitin Yadav, \\
    Mridul Sharma, 
    Ananya Kunisetty, 
    Saranya Darisipudi, 
    Nirmal Punjabi, \\
    Ganesh Ramakrishnan
    \affiliations
    Indian Institute of Technology Bombay \\
    \emails
    \small \texttt{23d1627@iitb.ac.in, npunjabi@iitb.ac.in, ganesh@cse.iitb.ac.in}
}
\begin{document}
\maketitle

\begin{abstract}
Large Language Models (LLMs) have shown remarkable capabilities for complex tasks, yet adaptation in medical domain, specifically mental health, poses specific challenges. Mental health is a rising concern globally with LLMs having large potential to help address the same. We highlight three primary challenges for LLMs in mental health - lack of high quality interpretable and knowledge grounded training data; training paradigms restricted to core capabilities, and evaluation of multi turn dialogue settings. 
Addressing it, we present oMind framework which includes training and aligning LLM agents for diverse capabilities including conversations; high quality $\sim$164k multi-task SFT dataset, as a result of our generation pipeline based on Structured Knowledge retrieval, LLM based pruning, and review actions. We also introduce oMind-Chat - a novel multi turn benchmark dataset with expert annotated turn level and conversation level rubrics. 
%oMind consistently outperforms it's base models
Our diverse experiments on both core capabilities and conversations shows oMind LLMs consistently outperform baselines. oMind-LLM also shows significantly better reasoning with up to 80\% win rate.

    %Mental health challenges affect nearly one in eight people globally, yet access to adequate diagnosis, prevention, and support remains limited. Recent advances in Large Language Models (LLMs) highlight their potential for symptom detection, dialogue generation, and personalized recommendations. However, existing adaptations of LLMs to the mental health domain are constrained by narrow task focus, limited clinical grounding, and poor interpretability, restricting their applicability in real-world scenarios. In this work, we introduce oMindLLM, a finetuned LLM for mental health with multi-task capabilities spanning question answering, classification, and conversational dialogue. To support this, we develop oMind-Fine, a large-scale dataset that enables robust learning across diverse downstream tasks, and medically grounded in ULMS KG triplets, and medical books. Further, we propose oMind-Converse, a novel benchmark for evaluating multi-turn conversational ability of LLMs in mental health contexts. Extensive experiments show that oMindLLM outperforms strong baselines in reasoning, interpretability, and dialogue alignment, while maintaining sensitivity critical for mental health applications. Our work provides the first integrated framework for building and evaluating clinically informed LLMs in mental health, laying a foundation for safe and effective deployment in real-world support settings. pipeline for generation - KGs, supportQA, targeted conversations, etc. DPO, ablations.
\end{abstract}
% \footnote{Code, models and dataset are available at \url{https://github.com/username/omind} and \url{https://huggingface.co/surajracha/oMind-LLM} \url{https://huggingface.co/datasets/surajracha/oMind-Chat/}, respectively.}
\section{Introduction}
Large Language Models (LLMs) have exhibited remarkable capabilities in handling complex tasks across a wide range of medical domains \cite{llm-2}, \cite{llm3}.
LLMs have strong potential to act as medical agent models for query assistance, diagnosis, and conversations.
However, applying these generalist models to medical domain, specifically to mental health, has remained a challenge \cite{llm-health}, \cite{llm-health2}. Responses to medical queries can suffer from incorrect or incomplete information, inconsistent correlations, or lack of interpretability \cite{llm-health1}. 

Mental health is a rapidly increasing concern globally, yet it poses a challenge to address it at scale \cite{mental-health2}. Tasks like chat supports, decision making, and query handling are of prime importance.
However, exploring adaptation of LLMs for mental health has been sparse. Further, the domain poses distinct challenges, like the need for rich domain specific data, multiple tasks, conversational abilities, as well as accurate interpretations \cite{mental-llm-1}. 

%Large Language Models (LLMs) have been increasingly used for medical applications like diagnosis, conversations, and recommendations. The field of mental health, especially user centric frameworks, can largely benefit from efficient adaptation of LLMs.

Previous strategies like RAG \cite{rag}, based on inference time retrieval with answering and simple Supervised Finetuning (SFT) are often limited for medical domain. RAG may lack the necessary correlations and detailed reasoning needed across knowledge \cite{rag1}, while off-the-shelf SFT often miss domain specific knowledge reasoning  \cite{sft1}. Further, they pose limitations like non-extendability to multiple task, lacking clinical grounding, lack of interpretability and conversational abilities, limiting the models real-world applicability.

%Previous strategies like RAG and simple SFT are limited - RAG lacks necessary correlations and detailed reasoning, while off-the-shelf SFT often misses domain-specific knowledge-intensive reasoning, clinical grounding, interpretability, and conversational abilities. Bridging LLMs for mental health requires strong domain specific understanding of common issues, ability for resourceful conversations, and correct reasoning.

This brings us to first bottleneck, namely lack of high quality interpretable training dataset for mental health.
The existing datasets are limited to a confined set of tasks judging only core capabilities (like classification) without including tasks like open ended QA and multi-turn chats \cite{llm-limit1}. The second bottleneck being, training models to effectively learn on multiple parallel tasks, especially a combination of both core capabilities and open ended chat based paradigms. 
User conversations requires special focus both for training and efficient evaluation. While QA and core knowledge evaluation has been largely explored, current studies are limited for detailed evaluation of multi turn dialogues. We highlight the third bottleneck as lack of efficient benchmarks for evaluation of multi turn dialogues in mental health. 
%This brings us to three key bottlenecks: (1) lack of high-quality interpretable training datasets—existing datasets focus on core capabilities (e.g., multiple choice) without open-ended QA and multi-turn chats; (2) training mod els to effectively learn on multiple parallel tasks, especially a combination of both core capabilities and open end chat based paradigms; and (3) lack of efficient benchmarks for multi-turn conversation evaluation in mental health, as current studies focus primarily on QA and core knowledge assessment.

%Previous works show LLM finetuning on mental health corpus. 

%While these methods are promising, they pose limitations like non-extendability to multiple task, lacking clinical grounding, lack of interpretability and conversational abilities, further limiting the models real work applicability. This brings us to first bottleneck, namely lack of high quality interpretable training dataset in this domain. We argue that existing datasets may be limited to a confined set of tasks judging only core capabilities without including open ended tasks.
Addressing these challenges, we introduce the oMind (\textbf{O}verall \textbf{Mind}) paradigm. oMind\footnotemark has three components: (i) oMind LLMs are mental health LLM agents, effectively finetuned for diverse interaction capabilities. (ii) oMind-SFT is a $\sim$164k, medically grounded instruction set using the UMLS knowledge graph and medical books grounding. (iii) oMind-Chat is novel multi-turn conversation benchmark with expert annotated rubrics.
\footnotetext{
Code: \url{https://github.com/surajrachaiitb/oMind},\\
Models: \url{https://huggingface.co/surajracha/oMind-Mistral/}, \url{https://huggingface.co/surajracha/oMind-Qwen/}, \url{https://huggingface.co/surajracha/oMind-Llama/},\\
Dataset: \url{https://huggingface.co/datasets/surajracha/oMind-Chat/}
}
%We seek to answer the question - how to adapt LLMs to mental health for wide range of tasks including chats and also enhance the quality of their inherent medical reasoning using knowledge graphs. 
In view of the highlighted challenges, we present our \textbf{Key Contributions} as follows.
\\
(i) We introduce oMind-LLMs, a range of specialized LLM agents for mental health grounded in medical knowledge. They are capable of seamless interactions across multiple tasks for both core capabilities like diagnostic classifications, MCQA, and open-ended capabilities including QA, support, and multi turn conversations. We also perform DPO based preference alignment to further enhance the model performances.
\\
(ii) To facilitate the above, we release oMind-SFT, a $\sim$164k multi-task, medically grounded instruction set for mental health finetuning curated from multiple open source datasets spanning over wide range of important topics. 
%The selection ensures the model learns both scientific knowledge as well as realistic situations in both mental health and psychology.
\\
(iii) To make the explanations medically accurate, we develop a generation framework involving retrieval action on UMLS knowledge graph triplets and medical books, followed by LLM based pruning, and post hoc NLI based review action. This ensures interpretable and accurate explanations. To address the challenge of moving beyond core capabilities, we use the MCQA and classification along with their explanations as seed data to generate supporting open ended QAs and dialogue conversations with targeted characteristics like symptom analysis, recommendations, etc. 
\\
(iv) Recognizing that standard benchmarks fail to capture the nuances of mental health dialogues, we introduce oMind-Chat, a novel benchmark with 961 instances, each containing multi turn (ranging 1-3 turns) related user queries, designed to assess alignment in mental health dialogues. Four domain expert annotate rubrics for turn wise focused response characteristics, as well as overall rubrics highlighting necessary information ideally required to be elicited throughout the conversation. This paradigm captures both turn level precision and overall coverage through multiple interactions. To our knowledge, oMind-Chat is first mental health benchmark which jointly evaluate turn-level and conversation-level coverage.

We conduct extensive experiments to evaluate the effectiveness of oMind across various of mental health tasks. Our results demonstrate that oMind-LLMs achieves substantial improvements over strong baselines in core tasks, reasoning, interpretability, and conversational quality, while maintaining sensitivity and reliability required in this domain. We perform various ablation studies including preference alignment and win rate analysis revealing further insights.

\section{Related Works}
\textbf{Mental Health LLMs. }  
Many domain specific Large Language Models have been recently developed for mental health applications. Prior efforts include finetuning discriminative models for domain adaptation, like Mental-BERT \cite{mentalbert}, Mental-RoBERTa \cite{mentalroberta}, MentalXLNet \cite{mentalxlnet}. MentaLLaMA \cite{mentalllama} marks the first extensive finetuning of LLaMA-2 models on mental health datasets like IMHI targeting classification along with explanations. Larger models like GPT-4 have shown promising capabilities for the same, however, the larger size and not be publicly available makes it challenging to use. IMHI \cite{mentalllama} is a unique benchmarking and training dataset which has multiple tasks based on a suite of existing datasets. It uses GPT to generate explanations and also slightly repurposed classification as a text generation task 
%\cite{xu2024mentalllm}. 
\\
\\
\textbf{Medical Conversations Evaluations.} 
Recent work has expanded into Q\&A and dialogue. PsyEval contains QA pairs to evaluate for emotional support skills \cite{psyeval}. PsyQA provides a large-scale Chinese dataset for generating long counseling text \cite{psyqa}. Recently, MentalChat 16k presented single turn mental health dialogues for training \cite{mentalchat16k}.
Although conversation datasets exist, for example -
SMILECHAT as multi-turn conversations from QA pairs by using ChatGPT to enhance dialogue diversity \cite{qiu2023smile}, ESConv \cite{liu2021towards}, AugESC \cite{zheng2022augesc} with emotional support dialogues for model training, ConvCounsel contains conversations strategies \cite{convcounsel2024},
they don't focus on benchmarking and evaluations. BiMediX \cite{bimedix} does MoE finetuning on Mixtral models with comparative evaluation focusing on GPT generated conversations from MCQA datasets. MT-Bench \cite{mtbench} and MT-Bench 101 \cite{mtbench-101} are one of the pioneers for chat based evaluations using LLM-as-a-judge in multi turn settings.
\\
\\
\textbf{Knowledge Grounded Medical Frameworks.} 
Many previous works have used structured knowledge sources like KGs for better answer generation. Think on Graphs (ToG) \cite{tog}, KGR \cite{kgr} for information retrieval and further reasoning. KG-SFT \cite{kg-sft} using KG retrieval followed by HIST algorithm pruning for medical SFT generation. KGARevion \cite{kgarevion} uses KG triplets for verification and generation of correct answer in response to a user query.

\begin{figure*}
    \centering
    \includegraphics[width=1\linewidth]{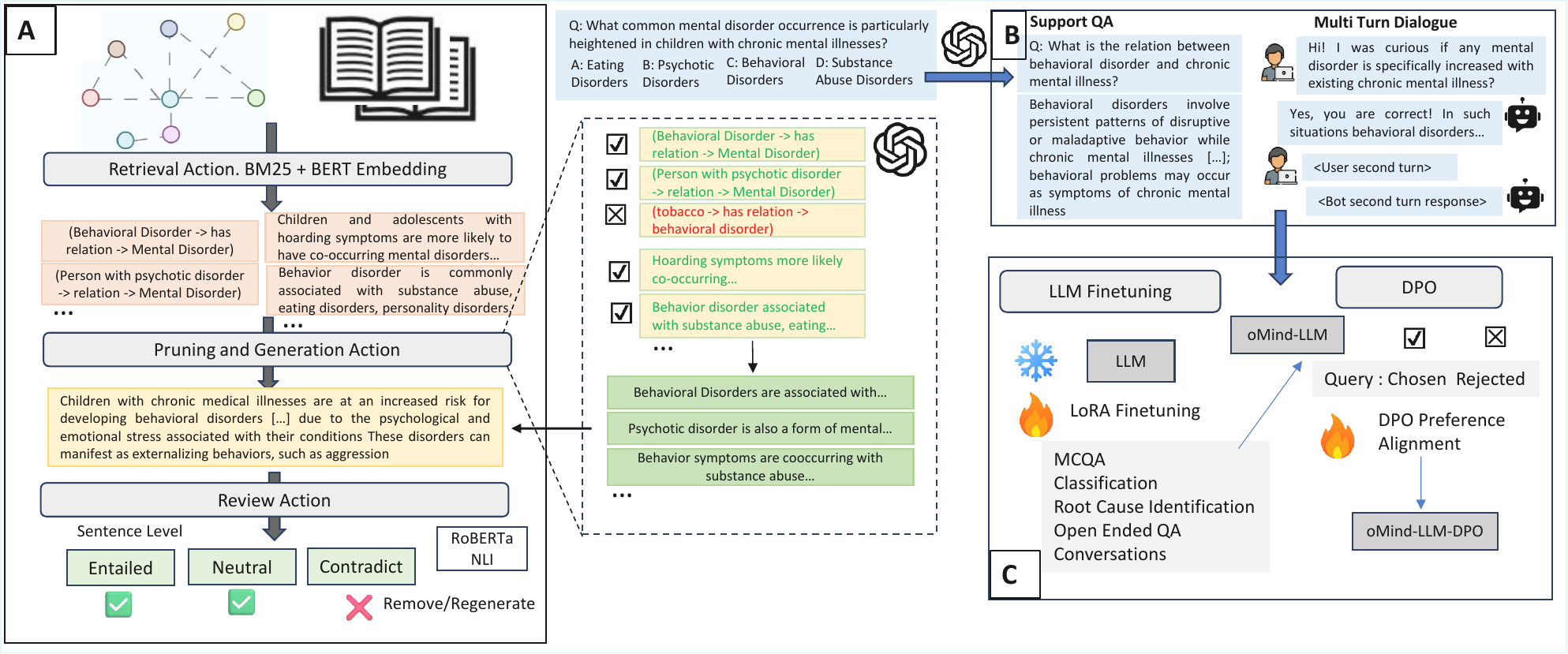}
    \caption{Complete pipeline for oMind SFT and model training. (A) Generation Pipeline consisting of retrieval stage, pruning \& generation stage, and review stage. (B) Generation of support QA and conversations using LLM. (c) LLM finetuning and preference learning.}
    \label{fig:framework}
\end{figure*}

\section{oMind - Methodology}
The overall approach towards oMind-LLM includes (1) Seed data collection, (2) Generation Framework, (3) Additional tasks, and (4) Preference Tuning (Figure \ref{fig:framework}).
%Our approach addresses key challenges for mental health LLM. 
We ground explanations for each QA pair in triplets from UMLS knowledge graph and standard psychology books (DSM-5 and ICD-11) to generate accurate reasoning and factual verification. We also extend to important open-ended tasks based on standard core knowledge like MCQA and Classification. The training includes LoRA finetuning and DPO to align model responses and avoid suboptimal reasoning.

\subsection{Seed Data Collection}
\label{sec:seed-data}
We start initially by collecting existing datasets called seed.
We directly collect mental health NLP datasets, while for general medical NLP datasets, restrict ourselves to domains related to mental health/psychology/psychiatry using predefined labels. Table \ref{tab:dataset-statistics} shows the statistics of overall datasets. We collect across four types - MCQA (multiple choice question answering), disorder classification, root cause classification, and conversations.
Overall, the topics covered range from anxiety, depression, OCD, trauma, counseling, psychology and basic psychiatry in terms of both factual and realistic situations. 
%The datasets contain two divisions: (i) mental health NLP datasets, and (ii) general medical NLP datasets with applied LLM based classification to filter mental health queries. While we directly use the data from (i), we perform a prompt based filtering on QA pairs to decide if the query relates to mental health knowledge or not for (ii). 
%Table <num> shows the details of overall datasets used along with their specific tasks and statistics. 

\subsection{Generation Framework}
The generation framework consists of (1) Retrieval Stage, (2) Pruning and Generation Stage, and (3) Review Stage.
\\
\\
\textbf{Retrieval Stage. } 
Consider an input query, $q$, and it's correct answer, $a$. Each query can be classified based on the format as (1) multiple choice consisting of query and a set of non-repeating options; (2) classification with input text as query and the classification labels; (3) long answer type with open ended question as the query.
%Consider an input query, $q$, and its known correct answer, $a$. We classify the query type largely into three categories based on its structuring and format - multiple choice option based where the query consists of the question and non repeating four options; classification based with the query consisting of a text and classifying labels similar across all queries; and long answer type where the query is an open ended question without options or labels. 
We use UMLS, a widely accepted standard medical knowledge graph and a set of standard open mental health and psychology books for the retrieval task. These act as the reference for medical grounding, accurate reasoning, and factual explanations. We combine BM25 (sparse) and embedding-based (dense) retrieval for better coverage. For each $q_i$, we retrieve triplet pairs from the KG and reference books paragraph chunks while eliminating any duplicate retrievals using both methods.
%For the retrieval task, we use UMLS knowledge graph (KG) and set of standard medical reference books based on psychology and mental health. See appendix <section num> for further details on the books used. These resources act as medical grounding for generating accurate reasoning and factual explanations. We use two retrieval methods, namely, BM25 for sparse and embedding similarity based for dense. For each $q_i$, we retrieve triplet pairs from the KG and paragraph chunks from the reference books set and eliminate any repeating retrievals. 
Finally, we have $\{T^i_1, T^i_2,..., T^i_n\}$ and $\{R^i_1, R^i_2,..., R^i_m\}$ where $T$ and $R$ are retrieved triplets and reference book chunks respectively for a given query $q_i$. In practice, both $n, m = 10$ with equal numbers from sparse and dense retrievals (five each). For MCQA task, the query consists of question along with the options, while for classification type, we perform an additional step - LLM based identification of key symptom phrases, thereby each text has list of symptom phrases and labels as queries followed by it's corresponding retrieval set.
\\
\\
\textbf{Pruning and Generation Stage. } 
The retrieved grounding set, being rule-based, is prone to contain irrelevant information. Pruning aims to mitigate redundant information and retain only query specific targeted knowledge. We use LLM (GPT-4o-mini here) as a pruning agent by leveraging its reviewing and reasoning capabilities. The LLM carefully reviews elements from set $T$ and $R$ against the input query, $q$, to identify only relevant elements that are related to and aid towards answering the query. To ease generation stage, the LLM turns selected triplets and book chunks into independent and complete sentences, conditioned strictly to output the same information without knowledge changes or additions. Post pruning, the grounding for $q_i$ becomes $G = \{g_1, g_2,..., g_n\}$ with each $g_{i}$ a grounding information.
%The above retrieved grounding pairs, being rule-based, can also contain irrelevant information. We aim to prune any irrelevant information and keep only targeted knowledge that helps answer the given query. 
%We use a LLM (GPT-4o-mini here) as a pruning agent. The LLM reviews both triplet pairs and book chunks to identify only the relevant knowledge and outputs independent information sentences based on it. We condition the LLM to strictly output the same information without additional knowledge from parametric data. Post pruning, the grounding for $q_i$ becomes $\{g_1, g_2,..., g_n\}$ where $g_{i}$ is a grounding information. Converting triplets into sentences gives an added advantage of easier usability for answer generation.
We pass the grounding set, along with query, $q$, and it's answer, $a$, into LLM (GPT-4o-mini) to generate a targeted explanation tailored to reach the given answer. For a richer explanation, we specifically prompt GPT to include elements like reasoning, factual explanation, and comparison, wherever applicable. GPT uses primarily the grounding information along with added parametric knowledge to fill in inconsistent spaces in generation. The grounding serves as a strong prior for explanations, as opposed to plain prompting. We call the generated explanation as $E$.
\\
\\
\textbf{Review Stage. } The reviewing stage ensures correctness in generation with reference to the grounding information and minimizes potential misaligned information. We use a NLI model, RoBERTa, to perform a sentence level analysis of generated explanation.
Specifically, we first separate $E$ into individual sentences $\{l_1, l_2,..., l_n\}$, while the grounding set is already a list of sentences. As matching all possible combinations requires large time, we use cosine similarity to find the top grounding information for each line, $l_i$. The pre-trained NLI model then classifies the pair into either of \textit{entailed, neutral, contradiction}. We remove those sentences which contradict the reference grounding above a given threshold (0.8). If more than two sentences contradict, we regenerate the explanation specifically targeting to resolve the contradicted information.

% \usepackage{booktabs}
% \usepackage{graphicx}
% \usepackage{booktabs}
% \usepackage{multirow}
% \usepackage{graphicx}
\begin{table}[]
\centering
\resizebox{0.7\columnwidth}{!}{%
\begin{tabular}{@{}llrr@{}}
\toprule
\textbf{Type} & \textbf{Dataset} & \textbf{\# Train} & \textbf{\# Test} \\ \midrule
\multirow{5}{*}{MCQA} & MHQA \cite{mhqa} & 25,000 & 500 \\
 & MedMCQA \cite{medmcqa} & 2,750 & 594 \\
 & MMLU-hsp \cite{mmlu} & 441 & 150 \\
 & MMLU-pp \cite{mmlu} & 462 & 150 \\
 & USMLE-Mental \cite{psyeval} & 498 & 219 \\ \midrule
\multirow{2}{*}{Disorder} & ANGST \cite{angst} & 6,990 & 500 \\
 & Dreaddit \cite{dreaddit} & 2,975 & 533 \\ \midrule
\multirow{2}{*}{Root Cause} & CAMS \cite{cams} & 1,666 & 290 \\
 & SAD \cite{sad} & 5,479 & 500 \\ \midrule
\multirow{3}{*}{Conversation} & ESConv \cite{esconv} & 910 & - \\
 & MentalChat 16k \cite{mentalchat16k} & 12,867 & - \\
 & CounselLMe \cite{counsellme} & 160 & - \\ \bottomrule
\end{tabular}%
}
\caption{Datasets statistics used for SFT}
\label{tab:dataset-statistics}
\end{table}

\subsection{Generating Further Tasks}
To add open-ended question answering and dialogue abilities during SFT, we leverage seed dataset limited to core tasks to generate additional long answer QA and conversations. This can also allows inter-task learning and better knowledge distillation.
The tasks along with their descriptions are as follows:
\\
\textbf{Supporting Long QA. } We define supporting long QA as long form question answer aimed at leveraging complex symptom relations, definitions, and core concepts found in MCQA and classification/root cause identification datasets. For each MCQA pair, GPT-4o-mini as LLM identifies important and non-trivial concepts, relations, and frames a question for the same. Similarly, for classification type, GPT identifies important symptom-disorder relations based on symptom phrases and forms long form questions for it. We restrict generation maximum up to three relevant support QAs for a given query. 
We follow the same generation framework for generating medically grounded answers for supporting QAs. Here the question becomes the query while the explanation itself is the answer.
In total, we generate $\sim$57k supporting long questions using the seed input. 
\\
\textbf{Multi-Turn Conversations. } Multi-turn conversations consist of multiple user-assistant dialogue pairs conditioned on previous responses. Apart from already existing chat datasets, we use MCQA and post classification datasets to generate synthetic multi turn targeted conversations. We define five medical categories of targeted conversations between a single user-assistant turn as follows: (i) symptom identification; (ii) disorder identification; (iii) follow-up questions; (iv) recommendations and solutions; (v) explanations and QA in conjunction with medical experts. Using GPT-4o-mini along with the query, $q$, answer, $a$, and the pruned medical grounding, we generate synthetic conversations with each turn primarily targeting one or more of the categories. The prompt also encourages to include mental health specific elements like support, affirmation, activate listening, validation, and proactive inquiry. Unlike existing datasets, the targeted approach ensures high quality and medically relevant chat based elements are explicitly present. The generated conversations range in between 3, 4, and 5 turns to include variations in format. We curate a total of $\sim$37.5k multi turn conversations based on the seed data and medical grounding.

\begin{figure}
    \centering
    \includegraphics[width=1\linewidth]{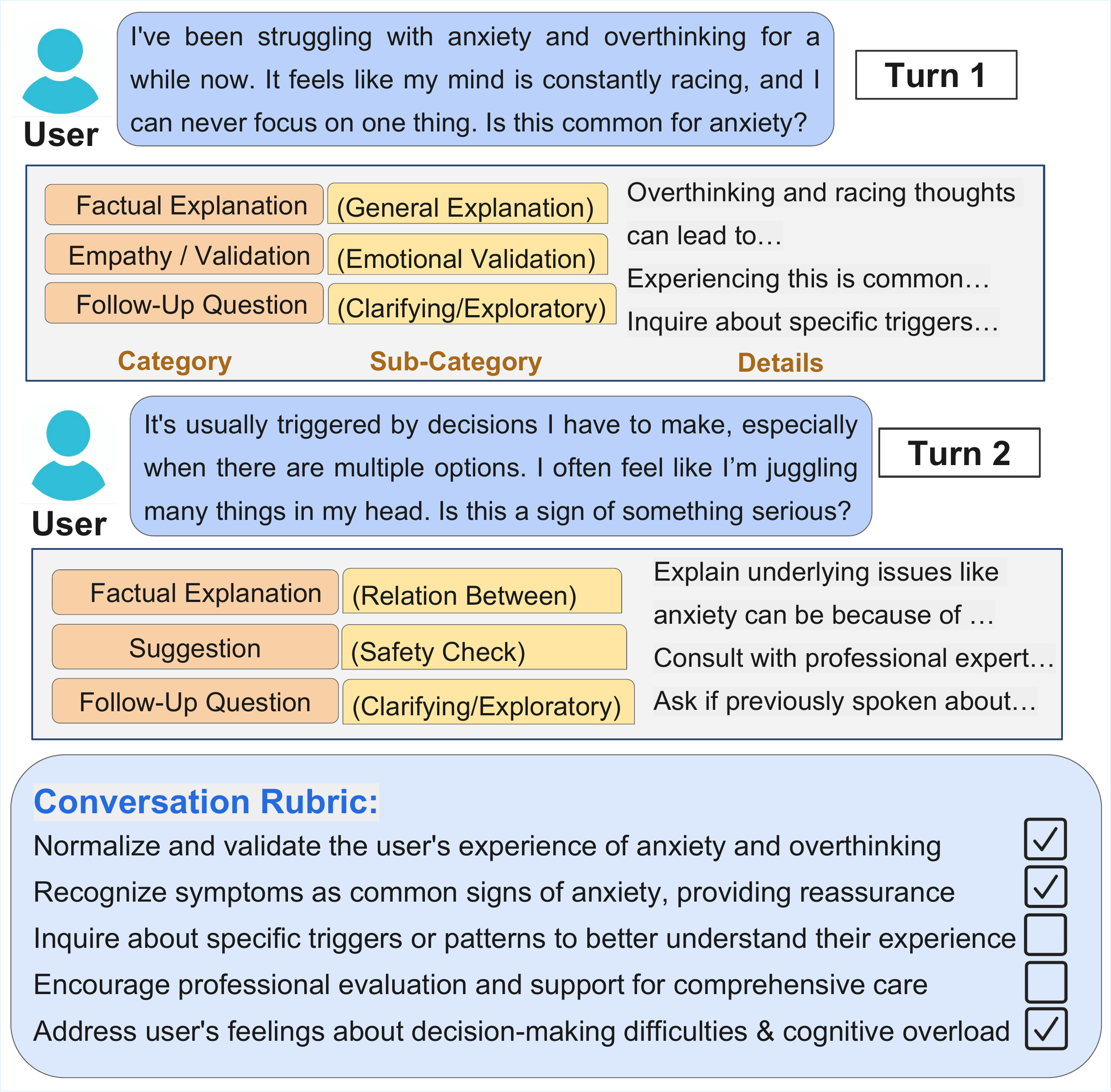}
    \caption{Illustration of oMind-Chat with rubrics.}
    \label{fig:omind-converse-fig}
\end{figure}

\subsection{SFT and Preference Tuning}
We generate overall $\sim$164k QA pairs forming the oMind-SFT dataset. We use LoRA based finetuning of LLMs, specifically training each model for 2 epochs, learning rate of $5 \times 10^{-5}$, with a weight decay of 0.01.
Each QA pair has instruction text, the input query, and grounded explanation, and correct answer.
Further on the SFT models, we perform preference alignment using DPO training over a subset of seen and unseen instances from oMind-SFT by curating a 20k preference dataset with 10k unseen instances (i.e. removed from SFT) while the rest being used in training.
We use GPT-as-a-judge for preferences. We pair between original answer-SFT model (5k instances), original answer-base model (5k instances), SFT model-base model (10k instances).
%GPT-based models are used generation, pruning, and preference labeling; while it can perform inference independent of it.

% Define pastel colors
\definecolor{pastelgreen}{RGB}{178, 228, 170}  % Soft mint green
\definecolor{pastelred}{RGB}{245, 255, 244}    % Soft pink/red

\begin{table*}[]
\centering
\resizebox{0.9\textwidth}{!}{%
\begin{tabular}{@{}llccccccccccc@{}}
\toprule
\multirow{2}{*}{\textbf{Model}} & \multirow{2}{*}{\textbf{Method}} & \multicolumn{5}{c}{\textbf{Classification}} & \multicolumn{6}{c}{\textbf{MCQA}} \\ \cmidrule(l){3-13} 
 &  & \multicolumn{1}{l}{\textbf{ANGST}} & \multicolumn{1}{l}{\textbf{dreaddit}} & \multicolumn{1}{l}{\textbf{CAMS}} & \multicolumn{1}{l}{\textbf{SAD}} & \multicolumn{1}{l|}{\textbf{AVG}} & \textbf{MHQA} & \textbf{MedMCQA} & \textbf{MMLU-hsp} & \textbf{MMLU-pp} & \textbf{USMLE} & \textbf{AVG} \\ \midrule
% ANGST: min=28.2, max=50.8
% dreaddit: min=44.1, max=78.2
% CAMS: min=13.5, max=42.8
% SAD: min=26.6, max=68.6
% Class AVG: min=29.3, max=57.2
% MHQA: min=53.8, max=81.2
% MedMCQA: min=52.4, max=85.5
% MMLU-hsp: min=61.8, max=89.9
% MMLU-pp: min=42.4, max=70.9
% USMLE: min=17.5, max=63.1
% MCQA AVG: min=48.0, max=76.3

\multicolumn{1}{l|}{\multirow{3}{*}{Llama 3.1 8B}} & \multicolumn{1}{l|}{Zero Shot} & \cellcolor{pastelgreen!12!pastelred}{31.0} & \cellcolor{pastelgreen!57!pastelred}{63.8} & \cellcolor{pastelgreen!58!pastelred}{30.7} & \cellcolor{pastelgreen!54!pastelred}{49.3} & \multicolumn{1}{r|}{\cellcolor{pastelgreen!52!pastelred}{43.7}} & \cellcolor{pastelgreen!35!pastelred}{63.6} & \cellcolor{pastelgreen!97!pastelred}{\underline{84.8}} & \cellcolor{pastelgreen!100!pastelred}{\textbf{89.9}} & \cellcolor{pastelgreen!71!pastelred}{62.8} & \cellcolor{pastelgreen!94!pastelred}{60.5} & \cellcolor{pastelgreen!85!pastelred}{\underline{72.3}} \\
\multicolumn{1}{l|}{} & \multicolumn{1}{l|}{LLM+KG} & \cellcolor{pastelgreen!31!pastelred}{35.4} & \cellcolor{pastelgreen!79!pastelred}{71.3} & \cellcolor{pastelgreen!69!pastelred}{33.8} & \cellcolor{pastelgreen!88!pastelred}{63.8} & \multicolumn{1}{r|}{\cellcolor{pastelgreen!78!pastelred}{51.1}} & \cellcolor{pastelgreen!78!pastelred}{75.4} & \cellcolor{pastelgreen!100!pastelred}{\textbf{85.5}} & \cellcolor{pastelgreen!95!pastelred}{\underline{88.5}} & \cellcolor{pastelgreen!93!pastelred}{\underline{69.0}} & \cellcolor{pastelgreen!100!pastelred}{\textbf{63.1}} & \cellcolor{pastelgreen!100!pastelred}{\textbf{76.3}} \\
\multicolumn{1}{l|}{} & \multicolumn{1}{l|}{LLM+Books} & \cellcolor{pastelgreen!49!pastelred}{39.4} & \cellcolor{pastelgreen!90!pastelred}{75.1} & \multicolumn{1}{r}{\cellcolor{pastelgreen!40!pastelred}{25.5}} & \cellcolor{pastelgreen!79!pastelred}{60.0} & \multicolumn{1}{r|}{\cellcolor{pastelgreen!74!pastelred}{50.0}} & \cellcolor{pastelgreen!59!pastelred}{70.2} & \cellcolor{pastelgreen!96!pastelred}{84.4} & \cellcolor{pastelgreen!76!pastelred}{83.3} & \cellcolor{pastelgreen!71!pastelred}{62.9} & \cellcolor{pastelgreen!38!pastelred}{34.9} & \cellcolor{pastelgreen!67!pastelred}{67.1} \\ \midrule

\multicolumn{1}{l|}{\multirow{3}{*}{MentaLLaMA 13B}} & \multicolumn{1}{l|}{Zero Shot} & \cellcolor{pastelgreen!54!pastelred}{40.7} & \cellcolor{pastelgreen!100!pastelred}{\underline{75.8}} & \cellcolor{pastelgreen!79!pastelred}{36.9} & \cellcolor{pastelgreen!88!pastelred}{63.7} & \multicolumn{1}{r|}{\cellcolor{pastelgreen!92!pastelred}{\underline{54.3}}} & \cellcolor{pastelgreen!34!pastelred}{63.2} & \cellcolor{pastelgreen!0!pastelred}{52.4} & \cellcolor{pastelgreen!50!pastelred}{75.9} & \cellcolor{pastelgreen!9!pastelred}{45.1} & \cellcolor{pastelgreen!27!pastelred}{30.1} & \cellcolor{pastelgreen!18!pastelred}{53.3} \\
\multicolumn{1}{l|}{} & \multicolumn{1}{l|}{LLM+KG} & \cellcolor{pastelgreen!47!pastelred}{39.2} & \cellcolor{pastelgreen!66!pastelred}{66.8} & \cellcolor{pastelgreen!43!pastelred}{26.2} & \cellcolor{pastelgreen!79!pastelred}{\underline{60.0}} & \multicolumn{1}{r|}{\cellcolor{pastelgreen!67!pastelred}{48.1}} & \cellcolor{pastelgreen!44!pastelred}{66.0} & \cellcolor{pastelgreen!20!pastelred}{59.2} & \cellcolor{pastelgreen!48!pastelred}{75.4} & \cellcolor{pastelgreen!16!pastelred}{47.1} & \cellcolor{pastelgreen!22!pastelred}{28.0} & \cellcolor{pastelgreen!25!pastelred}{55.2} \\
\multicolumn{1}{l|}{} & \multicolumn{1}{l|}{LLM+Books} & \cellcolor{pastelgreen!57!pastelred}{41.4} & \cellcolor{pastelgreen!61!pastelred}{65.1} & \cellcolor{pastelgreen!39!pastelred}{25.2} & \cellcolor{pastelgreen!68!pastelred}{55.4} & \multicolumn{1}{r|}{\cellcolor{pastelgreen!62!pastelred}{46.8}} & \cellcolor{pastelgreen!16!pastelred}{58.3} & \cellcolor{pastelgreen!5!pastelred}{54.2} & \cellcolor{pastelgreen!4!pastelred}{62.9} & \cellcolor{pastelgreen!0!pastelred}{42.4} & \cellcolor{pastelgreen!10!pastelred}{22.2} & \cellcolor{pastelgreen!0!pastelred}{48.0} \\ \midrule

\multicolumn{1}{l|}{\multirow{3}{*}{Mistral 7B}} & \multicolumn{1}{l|}{Zero Shot} & \cellcolor{pastelgreen!21!pastelred}{33.0} & \cellcolor{pastelgreen!62!pastelred}{65.5} & \cellcolor{pastelgreen!77!pastelred}{36.2} & \cellcolor{pastelgreen!73!pastelred}{57.4} & \multicolumn{1}{r|}{\cellcolor{pastelgreen!67!pastelred}{48.0}} & \cellcolor{pastelgreen!61!pastelred}{70.9} & \cellcolor{pastelgreen!49!pastelred}{68.7} & \cellcolor{pastelgreen!73!pastelred}{82.4} & \cellcolor{pastelgreen!44!pastelred}{55.2} & \cellcolor{pastelgreen!62!pastelred}{46.2} & \cellcolor{pastelgreen!58!pastelred}{64.7} \\
\multicolumn{1}{l|}{} & \multicolumn{1}{l|}{LLM+KG} & \cellcolor{pastelgreen!26!pastelred}{34.2} & \cellcolor{pastelgreen!55!pastelred}{62.9} & \cellcolor{pastelgreen!82!pastelred}{\underline{37.9}} & \cellcolor{pastelgreen!86!pastelred}{63.0} & \multicolumn{1}{r|}{\cellcolor{pastelgreen!72!pastelred}{49.5}} & \cellcolor{pastelgreen!84!pastelred}{77.1} & \cellcolor{pastelgreen!64!pastelred}{73.9} & \cellcolor{pastelgreen!61!pastelred}{79.2} & \cellcolor{pastelgreen!62!pastelred}{60.2} & \cellcolor{pastelgreen!66!pastelred}{47.8} & \cellcolor{pastelgreen!69!pastelred}{67.7} \\
\multicolumn{1}{l|}{} & \multicolumn{1}{l|}{LLM+Books} & \cellcolor{pastelgreen!38!pastelred}{37.0} & \cellcolor{pastelgreen!51!pastelred}{61.9} & \cellcolor{pastelgreen!58!pastelred}{30.7} & \cellcolor{pastelgreen!65!pastelred}{54.0} & \multicolumn{1}{r|}{\cellcolor{pastelgreen!59!pastelred}{45.9}} & \cellcolor{pastelgreen!48!pastelred}{67.1} & \cellcolor{pastelgreen!50!pastelred}{69.0} & \cellcolor{pastelgreen!79!pastelred}{84.0} & \cellcolor{pastelgreen!43!pastelred}{54.8} & \cellcolor{pastelgreen!53!pastelred}{41.9} & \cellcolor{pastelgreen!54!pastelred}{63.4} \\ \midrule

\multicolumn{1}{l|}{\multirow{3}{*}{Qwen 2.5 7B}} & \multicolumn{1}{l|}{Zero Shot} & \cellcolor{pastelgreen!43!pastelred}{38.1} & \cellcolor{pastelgreen!60!pastelred}{64.9} & \cellcolor{pastelgreen!22!pastelred}{20.0} & \cellcolor{pastelgreen!75!pastelred}{58.5} & \multicolumn{1}{r|}{\cellcolor{pastelgreen!57!pastelred}{45.4}} & \cellcolor{pastelgreen!54!pastelred}{68.8} & \cellcolor{pastelgreen!54!pastelred}{70.6} & \cellcolor{pastelgreen!86!pastelred}{86.2} & \cellcolor{pastelgreen!90!pastelred}{68.1} & \cellcolor{pastelgreen!55!pastelred}{42.8} & \cellcolor{pastelgreen!68!pastelred}{67.3} \\
\multicolumn{1}{l|}{} & \multicolumn{1}{l|}{LLM+KG} & \cellcolor{pastelgreen!45!pastelred}{38.6} & \cellcolor{pastelgreen!74!pastelred}{69.6} & \cellcolor{pastelgreen!73!pastelred}{35.2} & \cellcolor{pastelgreen!75!pastelred}{58.2} & \multicolumn{1}{r|}{\cellcolor{pastelgreen!75!pastelred}{50.4}} & \cellcolor{pastelgreen!66!pastelred}{72.1} & \cellcolor{pastelgreen!63!pastelred}{73.4} & \cellcolor{pastelgreen!67!pastelred}{80.8} & \cellcolor{pastelgreen!65!pastelred}{61.0} & \cellcolor{pastelgreen!64!pastelred}{47.2} & \cellcolor{pastelgreen!66!pastelred}{66.9} \\
\multicolumn{1}{l|}{} & \multicolumn{1}{l|}{LLM+Books} & \cellcolor{pastelgreen!57!pastelred}{41.4} & \cellcolor{pastelgreen!70!pastelred}{68.1} & \cellcolor{pastelgreen!37!pastelred}{24.5} & \cellcolor{pastelgreen!70!pastelred}{56.2} & \multicolumn{1}{r|}{\cellcolor{pastelgreen!65!pastelred}{47.6}} & \cellcolor{pastelgreen!60!pastelred}{70.4} & \cellcolor{pastelgreen!58!pastelred}{71.8} & \cellcolor{pastelgreen!81!pastelred}{84.6} & \cellcolor{pastelgreen!81!pastelred}{65.6} & \cellcolor{pastelgreen!0!pastelred}{17.5} & \cellcolor{pastelgreen!49!pastelred}{61.9} \\ \midrule

\multicolumn{1}{l|}{PMC-Llama 13B} & \multicolumn{1}{l|}{Zero Shot} & \cellcolor{pastelgreen!21!pastelred}{33.2} & \cellcolor{pastelgreen!0!pastelred}{44.1} & \cellcolor{pastelgreen!0!pastelred}{13.5} & \cellcolor{pastelgreen!0!pastelred}{26.6} & \multicolumn{1}{r|}{\cellcolor{pastelgreen!0!pastelred}{29.3}} & \cellcolor{pastelgreen!0!pastelred}{53.8} & \cellcolor{pastelgreen!50!pastelred}{69.0} & \cellcolor{pastelgreen!6!pastelred}{63.6} & \cellcolor{pastelgreen!21!pastelred}{48.5} & \cellcolor{pastelgreen!95!pastelred}{\underline{60.9}} & \cellcolor{pastelgreen!39!pastelred}{59.2} \\
\multicolumn{1}{l|}{MMedLM2 7B} & \multicolumn{1}{l|}{Zero Shot} & \cellcolor{pastelgreen!0!pastelred}{28.2} & \cellcolor{pastelgreen!13!pastelred}{48.8} & \cellcolor{pastelgreen!0!pastelred}{13.5} & \cellcolor{pastelgreen!0!pastelred}{26.8} & \multicolumn{1}{r|}{\cellcolor{pastelgreen!0!pastelred}{29.3}} & \cellcolor{pastelgreen!17!pastelred}{58.6} & \cellcolor{pastelgreen!27!pastelred}{61.5} & \cellcolor{pastelgreen!0!pastelred}{61.8} & \cellcolor{pastelgreen!35!pastelred}{52.5} & \cellcolor{pastelgreen!53!pastelred}{41.8} & \cellcolor{pastelgreen!25!pastelred}{55.2} \\ \midrule

\multicolumn{1}{l|}{oMind-Llama} & \multicolumn{1}{l|}{Zero Shot} & \cellcolor{pastelgreen!84!pastelred}{\underline{47.4}} & \cellcolor{pastelgreen!40!pastelred}{57.9} & \cellcolor{pastelgreen!55!pastelred}{29.7} & \cellcolor{pastelgreen!79!pastelred}{\underline{60.0}} & \multicolumn{1}{r|}{\cellcolor{pastelgreen!69!pastelred}{48.8 ($\uparrow$5.1)}} & \cellcolor{pastelgreen!77!pastelred}{75.0} & \cellcolor{pastelgreen!60!pastelred}{72.5} & \cellcolor{pastelgreen!29!pastelred}{70.1} & \cellcolor{pastelgreen!52!pastelred}{57.5} & \cellcolor{pastelgreen!49!pastelred}{40.2} & \cellcolor{pastelgreen!53!pastelred}{63.1 ($\downarrow$9.2)} \\
\multicolumn{1}{l|}{oMind-Qwen} & \multicolumn{1}{l|}{Zero Shot} & \cellcolor{pastelgreen!100!pastelred}{\textbf{50.8}} & \cellcolor{pastelgreen!15!pastelred}{49.3} & \cellcolor{pastelgreen!42!pastelred}{25.9} & \cellcolor{pastelgreen!23!pastelred}{36.4} & \multicolumn{1}{r|}{\cellcolor{pastelgreen!40!pastelred}{40.6 ($\downarrow$4.8)}} & \cellcolor{pastelgreen!86!pastelred}{\underline{77.6}} & \cellcolor{pastelgreen!63!pastelred}{73.7} & \cellcolor{pastelgreen!88!pastelred}{86.8} & \cellcolor{pastelgreen!100!pastelred}{\textbf{70.9}} & \cellcolor{pastelgreen!60!pastelred}{45.3} & \cellcolor{pastelgreen!81!pastelred}{70.9 ($\uparrow$3.6)} \\
\multicolumn{1}{l|}{oMind-Mistral} & \multicolumn{1}{l|}{Zero Shot} & \cellcolor{pastelgreen!57!pastelred}{41.2} & \cellcolor{pastelgreen!94!pastelred}{\textbf{76.2}} & \cellcolor{pastelgreen!100!pastelred}{\textbf{42.8}} & \cellcolor{pastelgreen!100!pastelred}{\textbf{68.6}} & \multicolumn{1}{r|}{\cellcolor{pastelgreen!100!pastelred}{\textbf{57.2} ($\uparrow$9.2)}} & \cellcolor{pastelgreen!100!pastelred}{\textbf{81.2}} & \cellcolor{pastelgreen!64!pastelred}{73.6} & \cellcolor{pastelgreen!83!pastelred}{85.2} & \cellcolor{pastelgreen!56!pastelred}{58.5} & \cellcolor{pastelgreen!48!pastelred}{39.6} & \cellcolor{pastelgreen!69!pastelred}{67.6 ($\uparrow$2.9)} \\ \bottomrule
\end{tabular}%
}
\caption{Performance comparison (\% F1 scores) of oMind models and baselines across Classification and MCQA tasks. AVG is the average score. Best and second best performance are in bold and underline respectively. We also report average \% gain/dip for oMind models compared with base models. LLM+KG and LLM+Books is baseline with retrieved KG and Books Chunks respectively.}
\label{tab:overall-results}
\end{table*}

% Define pastel yellow colors
\definecolor{pastelyellow}{RGB}{255, 237, 130}  % Pastel yellow
\definecolor{lightyellow}{RGB}{255, 255, 245}   % Nearly white yellow

\begin{table}[]
\centering
\resizebox{\columnwidth}{!}{%
\begin{tabular}{@{}lccc|ccc@{}}
\toprule
\multirow{2}{*}{\textbf{Model}} & \multicolumn{3}{c|}{\textbf{Turn Level}} & \multicolumn{3}{c}{\textbf{Conversation Level}} \\ \cmidrule(l){2-7} 
 & \textbf{Binary} & \textbf{Scale 0-10} & \textbf{Likert 1-5} & \textbf{Binary} & \textbf{Scale 0-10} & \textbf{Likert 1-5} \\ \midrule
% Turn Binary: min=0.36, max=0.73
% Turn Scale: min=2.8, max=6.0
% Turn Likert: min=2.6, max=4.0
% Conv Binary: min=0.45, max=0.79
% Conv Scale: min=3.4, max=6.5
% Conv Likert: min=2.8, max=4.1

\multicolumn{1}{l|}{Llama 3.1 8B} & \cellcolor{pastelyellow!75!lightyellow}{0.64} & \cellcolor{pastelyellow!75!lightyellow}{5.2} & \cellcolor{pastelyellow!78!lightyellow}{3.7} & \cellcolor{pastelyellow!91!lightyellow}{0.76} & \cellcolor{pastelyellow!90!lightyellow}{6.2} & \cellcolor{pastelyellow!84!lightyellow}{3.9} \\
\multicolumn{1}{l|}{MentaLLaMA 13B} & \cellcolor{pastelyellow!0!lightyellow}{0.36} & \cellcolor{pastelyellow!0!lightyellow}{2.8} & \cellcolor{pastelyellow!0!lightyellow}{2.6} & \cellcolor{pastelyellow!0!lightyellow}{0.45} & \cellcolor{pastelyellow!0!lightyellow}{3.4} & \cellcolor{pastelyellow!0!lightyellow}{2.8} \\
\multicolumn{1}{l|}{Mistral 7B} & \cellcolor{pastelyellow!67!lightyellow}{0.61} & \cellcolor{pastelyellow!65!lightyellow}{4.9} & \cellcolor{pastelyellow!78!lightyellow}{3.7} & \cellcolor{pastelyellow!91!lightyellow}{0.76} & \cellcolor{pastelyellow!83!lightyellow}{6.0} & \cellcolor{pastelyellow!84!lightyellow}{3.9} \\
\multicolumn{1}{l|}{Qwen 2.5 7B} & \cellcolor{pastelyellow!64!lightyellow}{0.60} & \cellcolor{pastelyellow!62!lightyellow}{4.8} & \cellcolor{pastelyellow!78!lightyellow}{3.7} & \cellcolor{pastelyellow!85!lightyellow}{0.74} & \cellcolor{pastelyellow!80!lightyellow}{5.9} & \cellcolor{pastelyellow!84!lightyellow}{3.9} \\
\multicolumn{1}{l|}{PMC-Llama 13B} & \cellcolor{pastelyellow!10!lightyellow}{0.40} & \cellcolor{pastelyellow!6!lightyellow}{3.0} & \cellcolor{pastelyellow!7!lightyellow}{2.7} & \cellcolor{pastelyellow!17!lightyellow}{0.51} & \cellcolor{pastelyellow!12!lightyellow}{3.8} & \cellcolor{pastelyellow!0!lightyellow}{2.8} \\
\multicolumn{1}{l|}{MMedLM2 7B} & \cellcolor{pastelyellow!45!lightyellow}{0.53} & \cellcolor{pastelyellow!37!lightyellow}{4.0} & \cellcolor{pastelyellow!42!lightyellow}{3.2} & \cellcolor{pastelyellow!58!lightyellow}{0.65} & \cellcolor{pastelyellow!48!lightyellow}{4.9} & \cellcolor{pastelyellow!38!lightyellow}{3.3} \\ \midrule
\multicolumn{1}{l|}{oMind-Llama} & \cellcolor{pastelyellow!54!lightyellow}{0.56} & \cellcolor{pastelyellow!53!lightyellow}{4.5} & \cellcolor{pastelyellow!50!lightyellow}{3.3} & \cellcolor{pastelyellow!55!lightyellow}{0.64} & \cellcolor{pastelyellow!51!lightyellow}{5.0} & \cellcolor{pastelyellow!53!lightyellow}{3.5} \\
\multicolumn{1}{l|}{oMind-Qwen} & \cellcolor{pastelyellow!100!lightyellow}{\textbf{0.73}} & \cellcolor{pastelyellow!100!lightyellow}{\textbf{6.0}} & \cellcolor{pastelyellow!100!lightyellow}{\textbf{4.0}} & \cellcolor{pastelyellow!100!lightyellow}{\textbf{0.79}} & \cellcolor{pastelyellow!100!lightyellow}{\textbf{6.5}} & \cellcolor{pastelyellow!100!lightyellow}{\textbf{4.1}} \\
\multicolumn{1}{l|}{oMind-Mistral} & \cellcolor{pastelyellow!72!lightyellow}{0.63} & \cellcolor{pastelyellow!71!lightyellow}{5.1} & \cellcolor{pastelyellow!78!lightyellow}{3.7} & \cellcolor{pastelyellow!82!lightyellow}{0.73} & \cellcolor{pastelyellow!77!lightyellow}{5.8} & \cellcolor{pastelyellow!84!lightyellow}{3.9} \\ \bottomrule
\end{tabular}%
}
\caption{Conversational performance on oMind-Chat benchmark. Models are evaluated at turn-level and conversation-level using three metrics: Binary rubric Coverage, 0-10 Scale, and Likert (1-5).}
\label{tab:conv-results}
\end{table}

\section{oMind-Chat Benchmark}
\label{sec:omind-chat}
We introduce oMind-Chat, a comprehensive benchmark dataset comprising of coherent user queries and specific rubrics in mental health and psychology contexts. Unlike traditional QA benchmarks, oMind-Chat assess multi-turn ability through open-ended dialogues paired with comprehensive evaluation rubrics. Each user turn is coherent and related to the previous query but also independent of previous response. We derive it from oMind-SFT, specifically sampling 1000 conversations comprising both synthetically generated and original (from other datasets) dialogues. We employ a two-level rubric generation approach using expert curated characteristics, structured LLM prompts, and post-annotation process. We present one such detailed example in Fig. \ref{fig:omind-converse-fig}.
\\
\\
\textbf{Turn-Level Rubrics. }  For each assistant response within a conversation turn, we use an expert evaluator prompt that instructs the LLM (GPT-4) to identify 2–5 key conversational elements that an ideal assistant response should contain. The prompt explicitly emphasizes generating prescriptive rubrics (what the response should include) rather than descriptive assessments (what it did include). The LLM selects from a predefined element inventory including: (1) Empathy/Validation, (2) Active Listening/Reflection, (3) Motivational Encouragement/Support, (4) Factual (5) Explanation/Psycho-education, (6) Follow-up Questions, (7) Diagnosis/Symptom Analysis, and (8) Suggestions/Recommendations.
Each element is annotated with a specific subtype (e.g., “Emotional Validation”) and a concrete short description of expected content, such as “Should acknowledge the user’s experience of anxiety and overthinking, normalizing these feelings to reduce stigma and foster understanding”.
\\
\textbf{Conversation-Level Rubrics. } The goal is to form an overall conversation rubrics aimed at presenting all necessary information that needs to be elicited by the chatbot during the User-LLM dialogues. Given the complete multi-turn conversation and the generated turn-level rubrics, the LLM infers 3–10 concise bullet points representing essential topics, insights, or clinical reasoning points that the assistant should cover across the entire dialogue. We use the pruned information, turn-level rubrics, and GPT parametric knowledge to draft the rubrics. These rubrics capture overarching therapeutic goals such as “Normalize and validate the user’s experience of anxiety and overthinking” or “Emphasize the benefits of combining medication with therapy for mental health improvement.” This approach ensures evaluation of both immediate response quality and longitudinal conversation coherence.
\\
\textbf{Annotation Process. } Four psychology experts (three primary and one secondary annotator) with a minimum of master’s qualifications in psychology review and validate the rubrics, ensuring accuracy and appropriateness. We employ a structured two-level evaluation protocol. The primary experts independently provide decisions for existing rubric points, additions for per-turn and conversation-level rubrics, and user query update. They can also reject a dataset instance if not valid or coherent. Each annotator is given same instructions without knowledge of other annotator details.
%We provide the complete details of the annotation schema in appendix <ref>.
The secondary expert's role is to consolidate the final decision for each rubric point. This includes resolving conflicting decisions for specific points between primary experts, and verifying and consolidating additional rubric points suggested.
\\
\textbf{Evaluation. } We introduce three scores: binary - check if given rubric point is directly present (0) or not (1) followed by average for each turn/conversation; score 0-10 - giving a score on scale of 0-10 for accurate presence of each rubric point followed by average; Likert 1-5 - Likert scale evaluation for overall comparison with complete rubric points.

We remain with 961 total conversations (535 single turn and 426 three turns), and a total 1,809 user-bot dialogue turns.

% \usepackage{booktabs}
% \usepackage{multirow}
% \usepackage{graphicx}
% \usepackage{booktabs}
% \usepackage{multirow}
% \usepackage{graphicx}

% \usepackage{booktabs}
% \usepackage{multirow}
% \usepackage{graphicx}

\begin{figure*}[t]
    \centering
    \includegraphics[width=1\linewidth]{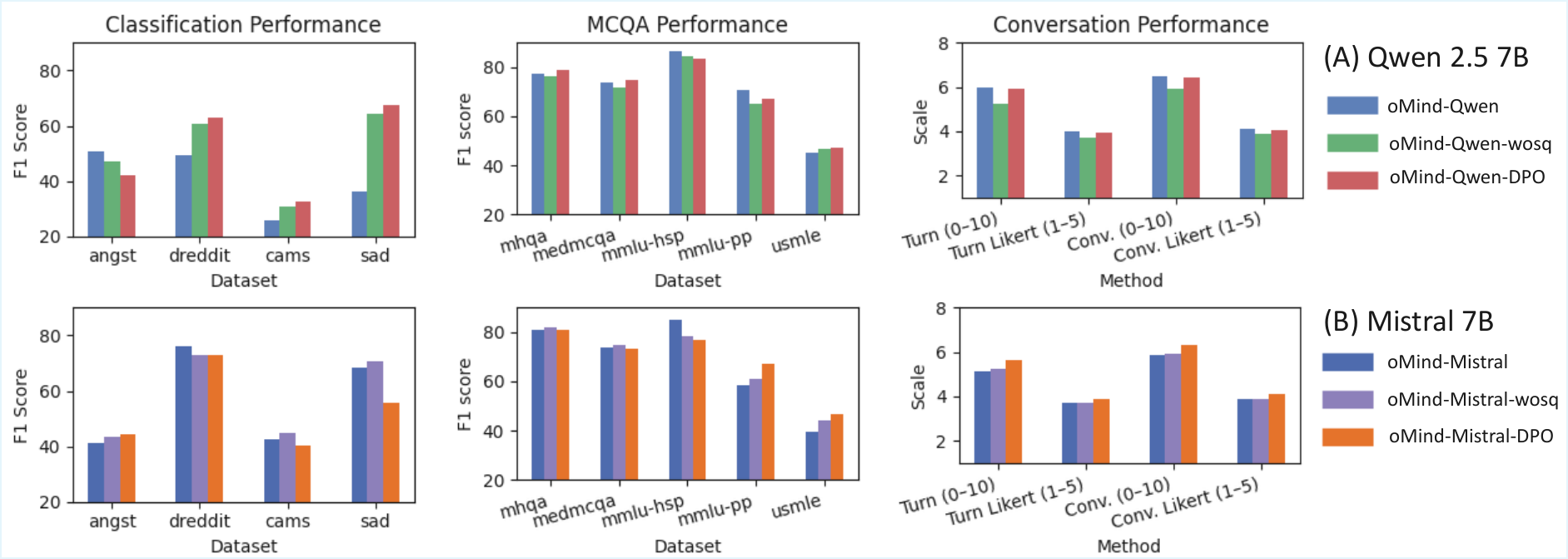}
    \caption{Impact of Supporting QA removal (wosq) and DPO alignment on Qwen-2.5-7B and Mistral-7B performance across classification, MCQA, and conversational tasks.}
    \label{fig:dpo-supportqa}
\end{figure*}

\section{Experimental Setup}
%This section outlines various experimental setups used to assess the framework, along with the datasets and baselines used.
\subsection{Datasets}
We use various mental health NLP datasets for evaluation.
We use the test split of all datasets as reported in table \ref{tab:dataset-statistics} for evaluation. The evaluation datasets are primarily of three categories, namely, MCQA, Classification (both disorder and root cause identification), and multi-turn chat based.

\subsection{Baselines}
We evaluate oMind-LLMs against three baselines. (i) \textit{Base LLM:} leveraging standalone LLMs for the given downstream tasks. (ii) \textit{LLM + KG:} wherein the LLM uses top-10 retrieved UMLS KG triplets as in-context knowledge for answering the given query. (iii) \textit{LLM + Books:} similar to previous baseline except the LLM uses retrieved medical books chunks as the in-context knowledge.
For model baselines, we use six models: instruct versions of Qwen-2.5-7B, Mistral-7B-v3, Llama-3.1-8B, along with medical LLMs including MentaLLaMA-13B, PMC-Llama 13B, MMedLM 7B.

\subsection{Experiments}
We span our experiments across multiple settings to allow detailed analysis.
\\
(i) We LoRA finetune three LLMs - Mistral-7B-v3, Qwen-2.5-7B, and Llama-3.1-8B on complete \~164k oMind-SFT dataset followed by DPO preference alignment on Qwen-2.5-7B and Mistral-7B-v3 to compare with SFT counterparts.
\\
(ii) We report F1 scores on multiple choice and classification type benchmarks and average scores assessing core knowledge capabilities across models and baselines.
%(ii) We perform DPO-based preference alignment by curating a 20k preference dataset with 10k unseen instances (i.e. removed from SFT) while the rest being used in training. The alignment comes from GPT-as-a-judge model and is paired between combinations of original answer, response by SFT model, and response by base model. Further details of the exact configurations used is provided in appendix \ref{}. 
\\
(iii) We baseline and report scores on oMind-Chat benchmark to evaluate multi-turn dialogue and long QA abilities following the evaluation strategy mentioned in section \ref{sec:omind-chat}. The previous turns are fed in-context as priors for next turn response.
\\
(iii) We perform ablation studies by finetuning Qwen-2.5-7B and Mistral-7B-v3 excluding Supporting QA to assess its effects on knowledge distillation; comparing with SFT and preference tuned models.
\\
(vi) To bring in interpretability and correctness of explanations, we report win rate scores between oMind-LLM (both SFT and DPO) and base LLM, using GPT-as-a-judge for those queries correctly answered (200 each). We also report explanation win rate scores across parameters between grounded explanation using the Generation Framework and GPT prompted explanations labelled by two annotators (100 instances). The parameters are: Factual Richness, Faithfulness, and Clarity.

% \section{Results \& Discussions}
% \textbf{Core Knowledge Tasks. }
% \\
% \textbf{Conversations. }

% \subsection{Effects of DPO \& Support QA}

% \section{Analysis}
% \subsection{Explanation Win Rate}
% \subsection{oMind and w/o win rate}

%Core
%Conversations
%Analysis
%Ablation
\section{Results \& Discussions}
\textbf{Core Capabilities. } Table \ref{tab:overall-results} reports F1 scores for MCQA and classification across all models. We find that oMind models consistently outperform their base counterparts in zero-shot settings in many of the datasets. For classification, except for oMind-Qwen which showed an average performance dip, both SFT Llama and Mistral showed a steep average performance increase of 5.1\% and 9.2\% respectively. Similarly, for MCQA, even though Llama showed an average strong performance dip, both Qwen and Mistral SFT showed performance increase. Considering classification and MCQA in separate lens, we see that SFT is more stable for classification. oMind-Mistral is the best classification model, leading 3/4 datasets with the highest average score (57.2\%). - indicating effective architectural alignment with our instruction-tuning strategy.
Even Llama shows modest performance gain with second-best model on two datasets, validating the effectiveness of our medically grounded training approach. While we also observe that Qwen-7B SFT results in decreased classification ability.
For individual datasets in MCQA, the paradigm remains tricky. Llama-3.1 8B unexpectedly outperforms on MMLU and USMLE, with performance comparable to even GPT-4. While for MHQA, the gain of 17.3\% F1 score using oMind is clearly evident. 
Excluding Llama-3.1 8B for the moment, for MCQA, oMind models again show an overall improvement in performance from their base models, except for Mistral with USMLE. 
Both retrieval baselines can improve base model performance. Generally, KG triplets helps increase capabilities, while books retrieval can tend to harm in some situations. Outperforming them indicates that oMind models have internalized domain knowledge. Medical-specific models (PMC-Llama, MMedLM2) generally lag behind, indicating that general medical knowledge alone is insufficient without targeted mental health adaptation.
Overall, oMind-Mistral shows best average performance for classification and if not considering Llama 3.1 8B, oMind-Qwen gains highest performance in MCQA.
\\
\textbf{Conversation Performance.} 
Table \ref{tab:conv-results} shows results on oMind-Chat benchmark across the three metrics: binary coverage, scaled scores (0--10), and Likert ratings (1--5) for turn level and conversation level evaluations. Overall, we observe that oMind-Qwen achieves the best performance on both turn level and conversation level. Looking at turn performances, we find that along with oMind-Qwen, oMind-Mistral shows modest gain over the base model with binary score increasing by 0.03 and 0.3 on 0--10 scale. However, Llama 3.1 8B base model itself shows significant chat capabilities, therefore finetuning actually hurting. 
Similarly, on a conversation level oMind-Qwen maintains its superiority (0.79 binary, 6.5 scaled, 4.1 Likert). oMind-Mistral shows slight decreased or constant (Likert) conversation-level performance (0.73 vs.\ 0.76) despite turn-level improvements, suggesting difficulty in maintaining coherent therapeutic goals over extended dialogues.
We also make an important observation that MentaLLaMA 13B which was finetuned on core mental health tasks over Llama 2 13B doesn't show strong conversation capabilities. Similarly, PMC-Llama 13B and MMedLM2 7B, although medical models are not strong conversation agents.

\subsection{Effects of DPO and Support QA}
Figure \ref{fig:dpo-supportqa} shows an important comparison between performance after preference alignment through DPO and an ablation study based on finetuning without support QA. Firstly, comparing the effects of preference tuning - DPO in general aid a lot for oMind-Qwen. Especially, the severe performance of oMind-Qwen on SAD is fixed after preference alignment with more than 25\% F1 score gain. Similarly for MCQA, DPO on oMind-Qwen improves performance for 3/5 datasets. Although, at certain points (eg. ANGST), DPO decreases performance, showing need to a deeper analysis. While DPO on oMind-Mistral shows a clear advantage for MCQA with gain for 4/5 datasets, DPO for classification adversely affects performance for dreaddit and SAD. The trends for SFT without support QA (wosq) vary significantly -- while there is a performance dip when support QAs are removed for Qwen MCQA and almost no significant difference for Mistral MCQA, the presence of support QA actually hurts classification for some datasets. However, the role of support QA becomes significant in conversations for best model oMind-Qwen, where there is a performance dip in absence of it. 
For conversational tasks, Qwen shows similar performance between base and DPO variants, both outperforming the Supporting-QA-removed setting, indicating that Supporting QA is crucial for conversational quality. Conversely, Mistral shows comparable performance between base and Supporting-QA-removed variants, while DPO improves outcomes, suggesting preference alignment is more critical than knowledge distillation for dialogue quality. These divergent patterns emphasize that optimal training strategies must be tailored to specific model architectures rather than applied universally.

%\textbf{Task-Specific Patterns.}
%Figure~2 reveals nuanced effects across tasks. In classification, Supporting QA provides minor improvements for both models, except on ANGST for Qwen and dreaddit for Mistral. DPO improves Qwen performance in most cases (except ANGST) but degrades Mistral performance (except ANGST). In MCQA, removing Supporting QA generally decreases Qwen performance (except USMLE) while improving Mistral performance (except MMLU-hsp), with no consistent trend observed for DPO.

%For conversational tasks, Qwen shows similar performance between base and DPO variants, both outperforming the Supporting-QA-removed setting, indicating that Supporting QA is crucial for conversational quality. Conversely, Mistral shows comparable performance between base and Supporting-QA-removed variants, while DPO improves outcomes, suggesting preference alignment is more critical than knowledge distillation for dialogue quality. These divergent patterns emphasize that optimal training strategies must be tailored to specific model architectures rather than applied universally.

\begin{table}[t]
\centering
\resizebox{0.9\columnwidth}{!}{%
\begin{tabular}{@{}l|rrr@{}}
\toprule
\textbf{Models} & \textbf{Win (\%)} & \textbf{Tie (\%)} & \textbf{Lose (\%)} \\
\midrule
oMind-Llama vs Llama               & \textbf{77.0} & 10.5  & 12.5 \\
oMind-Mistral vs Mistral           & \textbf{76.0} & 23.0  & 1.0  \\
oMind-Qwen vs Qwen                 & \textbf{80.0} & 4.0   & 16.0 \\
oMind-Mistral vs wosq              & 24.5  & \textbf{73.5} & 2.0  \\
oMind-Qwen vs wosq                 & 6.0  & \textbf{66.0} & 28.0 \\
oMind-Mistral-dpo vs oMind-Mistral & 14.0  & \textbf{71.5} & 14.5 \\
oMind-Mistral-dpo vs Mistral       & \textbf{87.0} & 1.5   & 11.5 \\
oMind-Qwen-dpo vs oMind-Qwen       & 22.0  & \textbf{36.5}  & 41.5 \\
oMind-Qwen-dpo vs Qwen             & \textbf{84.0} & 7.5  & 8.5 \\
\bottomrule
\end{tabular}%
}
\caption{Pairwise win-rate comparison for generated explanations.}
\label{tab:pairwise}
\end{table}

\begin{table}[h]
\centering
\resizebox{0.8\columnwidth}{!}{%
\begin{tabular}{@{}l|ccc@{}}
\toprule
 & \textbf{Factual Richness} & \textbf{Faithfulness} & \textbf{Clarity} \\
\midrule
oMind Win  & \textbf{0.67/ 0.58} & \textbf{0.49}/ 0.32 & \textbf{0.38}/ 0.31 \\
Tie    & 0.23/ 0.28 & 0.36/ \textbf{0.56} & 0.29/ 0.24 \\
GPT Win & 0.10/ 0.14 & 0.15/ 0.12 & 0.33/ \textbf{0.45} \\
\bottomrule
\end{tabular}%
}
\caption{Win Rate scores by annotator 1/annotator 2 for oMind and only GPT.}
\label{tab:kg-gpt-comparison}
\end{table}

\section{Win Rate Analysis.}
Table \ref{tab:pairwise} compares win rates ( \textit{i.e.}choosing better) for explanations. oMind models substantially outperform their base counterparts, with win rates of 80\% for oMind-Qwen, 76\% for oMind-Mistral, and 77\% for oMind-Llama. This validates the effectiveness of our SFT approach.
DPO-aligned models significantly outperform base models with oMind-Mistral-DPO having 87\% win rate and oMind-Qwen-DPO 84\% win rate, but tend to exhibit inconsistent improvements over their SFT oMind counterparts. oMind-Mistral-DPO shows marginal gains, while oMind-Qwen-DPO underperforms. 
The impact of removing Supporting QA varies significantly by model. Explanations from oMind-Qwen suffers substantial degradation without Supporting QA (6\% wins, 28\% losses compared to oMind-Qwen), demonstrating ineffective knowledge distillation. In contrast, Mistral shows a minimal impact (24.5\% wins, 73.5\% ties), which suggests it benefits more from direct task examples rather than auxiliary distillation. This again validates that the biggest contribution of preference tuning with improvement of conversation abilities.
Table \ref{tab:kg-gpt-comparison} compares similar win rates by two annotators on explanation with generation framework and direct GPT prompting. As observed, KG grounding adds significant Factual Richness with both annotators choosing it 67\% and 58\% times. Knowledge grounding also ensures faithfulness to the query, avoiding drift in topic. Although, due to the 'structured-ness' of KGs and books, clarity can sometimes be adversely affected -- with direct GPT explanations tending to be more smooth and clear in connecting between topics. The primary contribution of the generation framework is to add verified, medically grounded, rich explanations.

% \begin{table}[]
% \begin{tabular}{lrrr}
% models                             & \multicolumn{1}{l}{win} & \multicolumn{1}{l}{tie} & \multicolumn{1}{l}{loose} \\
% omind-llama vs llama               & 154                     & 21                      & 25                        \\
% oMind-Mistral vs mistral           & 152                     & 46                      & 2                         \\
% oMind-Qwen vs qwen                 & 160                     & 8                       & 32                        \\
% oMind-Mistral vs wosq              & 49                      & 147                     & 4                         \\
% oMind-Qwen vs wosq                 & 12                      & 132                     & 56                        \\
% oMind-Mistral-dpo vs oMind-Mistral & 28                      & 143                     & 29                        \\
% oMind-Mistral-dpo vs mistral       & 174                     & 3                       & 23                        \\
% oMind-Qwen-dpo vs oMind-Qwen       & 44                      & 73                      & 83                        \\
% oMind-Qwen-dpo vs qwen             & 168                     & 15                      & 17                       
% \end{tabular}
% \end{table}
\section{Conclusion}
We present oMind for mental health LLMs consisting of finetuned and preference aligned oMind-LLMs, knowledge grounded SFT, and oMind-Chat multi-turn conversation benchmark. By combining structured medical knowledge in explanation with multi task SFT, oMind shows improvements on downstream tasks and dialogue performance. The framework also significantly improves the reasoning ability over base models, making it medically accurate. 
Further results confirm the importance of preference tuning and support QA (open-ended) SFT especially for conversations.

%We present oMind, a knowledge-grounded multi-task framework for mental health LLMs with strong reasoning and conversational abilities. By combining structured medical knowledge retrieval and multi-task instruction tuning, oMind consistently improves classification, MCQA, and multi-turn dialogue performance. Our oMind-SFT dataset and oMind-Chat benchmark address key gaps in training and evaluation beyond single-turn paradigms, offering a scalable and interpretable approach for advancing mental health LLM research.

%We present oMind, a comprehensive framework for developing mental health LLMs that addresses critical limitations in clinical grounding, interpretability, and multi-task capabilities. Our medically grounded generation framework, leveraging UMLS knowledge graphs and psychology textbooks with NLI validation, enables oMindLLM to provide interpretable explanations while maintaining clinical accuracy. 

%The oMind-SFT dataset (164k samples) and oMind-Converse benchmark establish rigorous standards for training and evaluating mental health dialogue systems. Extensive experiments demonstrate that oMind models substantially outperform baselines across classification, MCQA, and conversational tasks, with architecture-specific insights revealing the importance of tailored training strategies.
\appendix

\section*{Ethical Statement}
%There are no ethical issues.
oMind is presented as a research work to study various results across evaluations. Even as research, they are designed as support systems for decisions. It is not designed to replace professional experts or for making stand alone deployment in real world settings.

%\section*{Acknowledgments}

\bibliographystyle{named}
\bibliography{ijcai26}

\end{document}